# Scalable and Interpretable Scientific Discovery via Sparse Variational Gaussian Process Kolmogorov-Arnold Networks (SVGP KAN)

Y. S. Ju

Department of Mechanical and Aerospace Engineering,

UCLA, Los Angeles, CA 90095, U.S.A.

**Abstract**

Kolmogorov-Arnold Networks (KANs) offer a promising alternative to Multi-Layer Perceptron (MLP) by placing learnable univariate functions on network edges, enhancing interpretability. However, standard KANs lack probabilistic outputs, limiting their utility in applications requiring uncertainty quantification. While recent Gaussian Process (GP) extensions to KANs address this, they utilize exact inference methods that scale cubically with data size *N*, restricting their application to smaller datasets. We introduce the Sparse Variational GP-KAN (SVGP-KAN), an architecture that integrates sparse variational inference with the KAN topology. By employing $M$ inducing points and analytic moment matching, our method reduces computational complexity from $O(N^3)$ to $O(NM^2)$ or linear in sample size, enabling the application of probabilistic KANs to larger scientific datasets. Furthermore, we demonstrate that integrating a permutation-based importance analysis enables the network to function as a framework for structural identification, identifying relevant inputs and classifying functional relationships.

## 1. Introduction

The intersection of deep learning and scientific discovery requires models that provide both accuracy and interpretability. The Kolmogorov-Arnold Network (KAN), introduced by Liu et al.[1,2], presents an alternative to traditional Multi-Layer Perceptrons (MLPs). Unlike MLPs, which fix activation functions on nodes, KANs place learnable univariate functions on edges. This structure aligns with the Kolmogorov-Arnold representation theorem, which states that multivariate continuous functions can be represented as finite compositions of univariate functions. While KANs are effective for symbolic regression and function approximation, they are deterministic and do not inherently provide uncertainty estimates. This can be problematic in fields such as physics, biology, and engineering, where models not only provide predictive outcomes but also reveal underlying physical laws and relationships. Furthermore, quantifying epistemic uncertainty is important for safety and decision-making.

Gaussian Processes (GPs) provide a Bayesian framework for non-parametric function approximation with principled uncertainty estimates. Combining the structural interpretability of KANs with the probabilistic properties of GPs presents a potential solution. Recent efforts have attempted to address this. A previous study[3] proposed Bayesian-KANs, which place probability distributions over the coefficients of B-splines. This captures weight-space uncertainty but does not fully model the function-space uncertainty inherent in non-parametric modeling. A more recent study[4] introduced GP-KANs, where edge functions are drawn directly from GP priors. However, their formulation relies on exact GP inference, which entails inverting a covariance matrix of size $N \times N$, leading to $O(N^3)$ computational complexity.

In the present work, we propose the Sparse Variational GP-KAN (SVGP-KAN) to help bridge the gap between interpretability, uncertainty quantification, and scalability. We leverage the framework of Sparse Variational Gaussian Processes[5,6] to approximate the posterior distribution using a set of learnable inducing points. This approach reduces the computational burden to $O(NM^2)$ overall, and $O(BM^2)$ per mini batch of size $B$, where $M \ll N$ is the number of inducing points. Additionally, we implement a dense matrix-based formulation of KAN networks inspired by a previous work[7], which can utilize GPU parallelism to train thousands of univariate GPs simultaneously. Finally, we incorporate explicit observation noise modeling and permutation importance analysis to distinguish signal from noise in physical systems.

Our contributions are fourfold:
- **Topology:** We integrate univariate GP edges within the KAN architecture, enabling factorized sparse variational inference across edges.
- **Uncertainty propagation:** We adapt analytic moment matching to KAN's additive edge aggregation, yielding closed-form layer means and variances.
- **Discovery mechanism:** We combine variational training with post-hoc permutation importance to robustly identify relevant features and functional forms.
- **Scalability:** We provide a batched matrix formulation that parallelizes univariate GP edges on GPUs.

## 2. Mathematical framework

The premise of this work is that the univariate functions $\phi$ in a KAN layer can be modeled as independent Gaussian Processes, and that the propagation of uncertainty through these layers can be computed using variational approximations. We denote the number of training samples as $N$, the number of inducing points as $M$, and the batch size as $B$.

### 2.1 Probabilistic topology

A standard KAN layer maps an input vector $\mathbf{x} \in \mathbb{R}^{D_{in}}$ to an output $\mathbf{y} \in \mathbb{R}^{D_{out}}$ via a matrix of univariate functions $\Phi = \{\phi_{j,i}\}$, where the $j$-th output is the sum of activations

$$y_j = \sum_{i=1}^{D_{in}} \phi_{j,i}(x_i).$$

In the SVGP-KAN, each function $\phi_{j,i}$ is modeled as an independent Gaussian Process with a zero-mean prior and a kernel function $k(\cdot,\cdot)$. Unlike standard Deep GPs which often involve multivariate mappings, the KAN topology enforces univariate edges. This simplifies the covariance structure, allowing the factorization of the variational distribution across edges.

### 2.2 Sparse variational inference

To improve scalability, we employ variational inference rather than exact marginalization. For each edge connecting input $i$ to output $j$, we introduce a set of $M$ inducing inputs $\mathbf{Z}_{ij} = \{z_1, \dots, z_M\}$ and inducing variables $\mathbf{u}_{ij} = f(\mathbf{Z}_{ij})$. We approximate the true posterior $p(f|\mathbf{y})$ with a variational distribution $q(f)$. Following the existing framework[6], we parametrize $q(\mathbf{u}_{ij})$ as a multivariate Gaussian $\mathcal{N}(\mathbf{m}_{ij}, \mathbf{S}_{ij})$, where $\mathbf{m}$ and $\mathbf{S}$ are learnable variational parameters. The conditional distribution $q(f)$ is obtained by marginalizing out the inducing variables. This formulation allows for the computation of the Evidence Lower Bound (ELBO) using mini batches of data, enabling stochastic gradient descent optimization. This decouples the computational cost from the total dataset size $N$, achieving $O(M^2)$ complexity per sample.

### 2.3 Analytic moment matching

A challenge in Deep GPs is propagating probability distributions through non-linear kernels. Since the output of a GP layer is a random variable, the input to the subsequent layer is stochastic. We employ Analytic Moment Matching[8] to approximate the marginal likelihood. Given an input distribution $x \sim \mathcal{N}(\mu, \sigma^2)$, we estimate the first two moments of the output $f(x)$. For the Squared Exponential (RBF) kernel with signal variance $\sigma_f^2$ and lengthscale $\ell$, the expected value of the kernel vector $\psi_1 = \mathbb{E}_x[k(x, \mathbf{Z})]$ is derived analytically:

$$[\psi_1]_m = \sigma_f^2 \left(\frac{\ell^2}{\ell^2 + \sigma^2}\right)^{1/2} \exp\left(-\frac{(z_m - \mu)^2}{2(\ell^2 + \sigma^2)}\right).$$

This equation demonstrates a regularization property of the architecture. The input variance $\sigma^2$ increases the squared lengthscale in the denominator: $\ell^2 \to \ell^2 + \sigma^2$. This implies that as the uncertainty about the input location increases, the effective lengthscale increases, smoothing the function. We compute the output mean and variance of the layer by aggregating these moments via additive composition across the KAN summation structure. Assumption of independence between edge GPs allows summing their variances, a commonly adopted mean-field approximation that simplified computation while offering robustness.

## 3. Structural identification via permutation importance

While standard KANs often rely on symbolic regression libraries to simplify learned functions, we employ a robust permutation-based importance metric to identify relevant features. After training the SVGP-KAN with a physics-informed noise lower bound to prevent overfitting to aleatoric noise, we evaluate feature importance by randomly permuting each input feature in the test set and measuring the increase in Mean Squared Error (MSE). Features that result in a negligible increase in MSE are classified as noise and pruned. This method is model-agnostic and tends to be robust to non-linear dependencies.

Furthermore, we classify functional relationships based on the learned posterior lengthscales $\ell$ and the visualized shape of the univariate functions. A learned lengthscale $\ell \to \infty$ implies that the RBF kernel has linearized locally to approximate a linear trend, whereas $\ell \ll 1$ implies high-frequency non-linear behavior. This allows the SVGP-KAN to categorize inputs without manual intervention.

## 4. Experiments and results

To evaluate the architecture's capabilities in structure identification and uncertainty quantification, we conducted three controlled experiments: two on synthetic datasets simulating noisy physical measurements and one on a standard regression benchmark.

**4.1 Basic discovery: feature selection and structural identification**

In the first experiment, we evaluated the model's ability to separate signal from noise and identify functional forms. The dataset consisted of $N = 600$ samples generated from a composite function containing a periodic signal, a linear trend, and a noise channel: $y = \sin(3\pi x_1) + 1.5x_2 + \varepsilon$ where $\varepsilon \sim \mathcal{N}(0, 0.1)$. The model was initialized with universal RBF kernels and $M = 20$ inducing points.

Our results (Figure 1) demonstrate three distinct behaviors. First, for the periodic input, the model learned a short kernel lengthscale, allowing the mean function to approximate the sinusoidal pattern. Second, for the linear input, the optimization process drove the lengthscale hyperparameter to large values ($\ell > 3.0$), allowing the probabilistic network to model linear trends without an explicit linear kernel prior. Third, for the noise channel, the permutation importance analysis indicated negligible contribution to the predictive accuracy, correctly identifying the feature as irrelevant.

### 4.2 Surface reconstruction

In the second experiment, we assessed the model's ability to reconstruct a continuous 2D manifold from noisy samples and quantify extrapolation uncertainty. The target function was a superposition of orthogonal waves, creating a "checkerboard" topology. The training data was sampled from the domain $x \in [-1,1]$, and we evaluated the model on an extended domain $x \in [-1.5,1.5]$ to test extrapolation behavior.

The SVGP-KAN reconstructed the underlying manifold, filtering out observation noise. The additive structure of the KAN topology captured the independent contributions of the sine and cosine components (Figure 2). The uncertainty quantification exhibited a pattern corresponding to the additive variance, suggesting that the analytic moment matching propagates input uncertainty.

### 4.3 Benchmark on Friedman regression

To validate the model on a standard regression task involving variable interactions and redundant features, we applied SVGP-KAN to the Friedman #1 dataset. This dataset

$$y = 10\sin(\pi x_0 x_1) + 20(x_2 - 0.5)^2 + 10x_3 + 5x_4 + \epsilon$$

requires models to capture non-linear interactions while ignoring 5 out of 10 input dimensions. We employed a permutation importance strategy post-training to robustly identify relevant features.

A 2-layer SVGP-KAN (10 → 10 → 1) was trained for 1500 epochs. To account for stochastic variability in initialization and noise generation, we conducted 5 independent trials. The model achieved a mean Test RMSE of $0.36 \pm 0.05$. The structural discovery capabilities were evaluated by tracking the selection rate of each feature across trials. The five true features ($x_0$ to $x_4$) were identified as relevant in 100% of trials. Conversely, the five noise features ($x_5$ to $x_9$) exhibited low stability, with selection rates ranging from 0% to 40%.

Visual diagnostics (Figure 3 top) confirm a clear separation in importance scores between signal and noise features. Furthermore, an analysis of the learned interaction surface (Figure 3 bottom) shows that the model captured, at least approximately, the non-linear saddle shape characteristic of the $x_0 x_1$ term, despite the additive constraints of the KAN topology. This suggests that deep SVGP-KANs can capture multiplicative interactions through composition, while the probabilistic framework retains feature selection capabilities in higher dimensions.

## 5. Conclusion

The SVGP-KAN combines the interpretability of Kolmogorov-Arnold Networks with the scalability and uncertainty quantification of Sparse Gaussian Processes. By using sparse variational inference instead of exact inference, we reduce the computational complexity from cubic to linear in sample size, enabling the use of probabilistic KANs on larger scientific datasets. The resulting architecture serves as a tool for structure identification, capable of determining functional forms and relevant variables in high-dimensional data. Limitations include the current restriction to additive structures, which requires additional network depth to model multiplicative interactions, and the choice of kernels. Future work will focus on extending this framework to multi-output regression tasks, integrating non-stationary kernels for discontinuous phenomena, and exploring multiplication layers.

**Code availability:** The code library and example test scripts are available at https://github.com/sungjuGit/svgp-kan

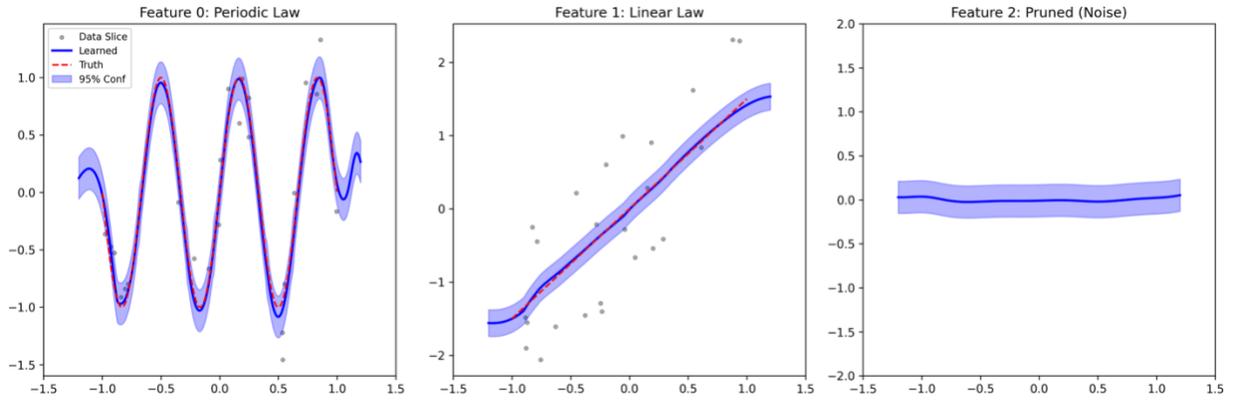

Figure 1: Results from the basic discovery experiment.

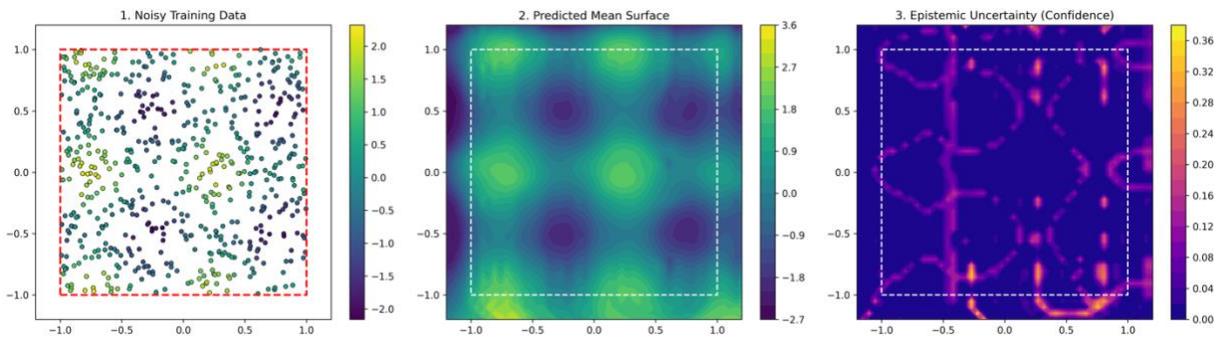

Figure 2: Results from the surface reconstruction experiment.

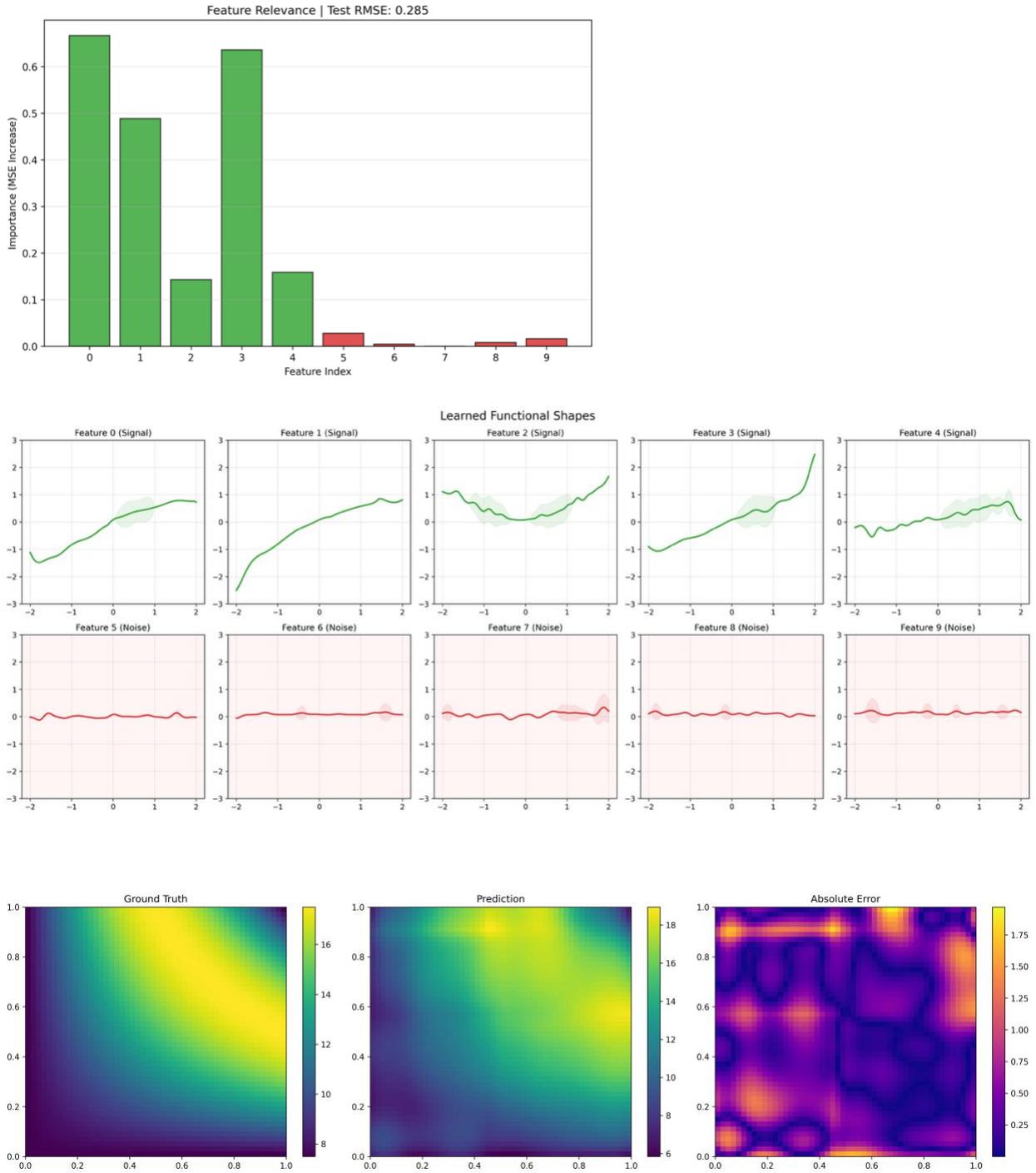

Figure 3: Results from the Friedman #1 benchmark tests.